\documentclass[letterpaper]{article} 
\usepackage{aaai2026}  
\usepackage{times}  
\usepackage{helvet}  
\usepackage{courier}  
\usepackage[hyphens]{url}  
\usepackage{graphicx} 
\usepackage{natbib}  
\usepackage{caption} 
\usepackage{algorithm}
\usepackage{algorithmic}

\usepackage{amsmath}
\usepackage{amssymb}
\usepackage{amsfonts}
\usepackage{mathrsfs}
\usepackage{textcomp}
\usepackage{xcolor}
\definecolor{crimson}{RGB}{220,20,60}
\definecolor{navy}{RGB}{65,105,225}
\usepackage{multirow}
\usepackage{verbatim}
\usepackage{enumitem}
\usepackage{booktabs}
\usepackage{subcaption}

\begin{document}
\title{Deep Incomplete Multi-View Clustering via Hierarchical Imputation and Alignment}

\author{
  Yiming Du, Ziyu Wang, Jian Li, Rui Ning, Lusi Li\thanks{Corresponding Author}\\
}

\affiliations{
    Department of Computer Science, Old Dominion University, Norfolk, VA 23529, USA\\
    \{ydu002, zwang007, jli038\}@odu.edu, \{rning, lusili\}@cs.odu.edu

}

\maketitle
\begin{abstract}
\begin{quote}
Incomplete multi-view clustering (IMVC) aims to discover shared cluster structures from multi-view data with partial observations. The core challenges lie in accurately imputing missing views without introducing bias, while maintaining semantic consistency across views and compactness within clusters. 
To address these challenges, we propose DIMVC-HIA, a novel deep IMVC framework that integrates hierarchical imputation and alignment with four key components: 
(1) view-specific autoencoders for latent feature extraction, coupled with a view-shared clustering predictor to produce soft cluster assignments; 
(2)  a hierarchical imputation module that first estimates missing cluster assignments based on cross-view contrastive similarity, and then reconstructs missing features using intra-view, intra-cluster statistics; 
(3) an energy-based semantic alignment module, which promotes intra-cluster compactness by minimizing energy variance around low-energy cluster anchors;
and (4) a contrastive assignment alignment module, which enhances cross-view consistency and encourages confident, well-separated cluster predictions. 
Experiments on benchmarks demonstrate that our framework achieves superior performance under varying levels of missingness.

\end{quote}
\end{abstract}

\begin{links}
    \link{Code}{https://github.com/YMBest/DIMVC-HIA}
\end{links}

\section{Introduction}

Multi-view clustering (MVC) has emerged as a fundamental paradigm in unsupervised learning, with broad applications in multimedia analysis~\cite{cao2025cross}, bioinformatics~\cite{zang2025rethinking}, and social network mining~\cite{zeng2024adversarial}. In real-world data representation, information is often collected from multiple heterogeneous sources--such as different sensors, modalities, or feature extractors--resulting in multi-view data that provides complementary representations of the same underlying phenomena. By integrating these diverse perspectives, MVC offers richer insights than single-view approaches~\cite{lu2024decoupled, guan2025structure}, enabling more robust pattern discovery, improved generalization, and a deeper understanding of complex data structures. 

Despite these advantages, conventional MVC methods rely on a critical yet often unrealistic assumption that all views are complete and fully observable \cite{chao2024incomplete}. Missing views are pervasive in real-world scenarios, arising from sensor failures, data corruption, transmission errors, or preprocessing artifacts. This incompleteness severely limits the applicability of  MVC techniques. To address this gap, incomplete MVC (IMVC) has emerged as an active research frontier, aiming to perform reliable clustering when some views are partially missing--sometimes leaving only one available view. The primary challenge of IMVC is to leverage available information effectively while mitigating the adverse effects of missing data, ensuring that clustering reliability and accuracy remain intact.

Recent advancements in IMVC have led to the development of various methodologies to handle missing views, including graph-based \cite{tan2024effective, lu2024decoupled}, matrix factorization-based~\cite{park2025smmf}, kernel-based~\cite{liu2023contrastive, ding2024multi}, subspace-based~\cite{ji2025anchors, gu2024noodle}, and deep learning-based techniques~\cite{du2025pgformer, wang2025energy, dong2025selective}. These methods generally fall into two categories. Imputation-based approaches~\cite{pu2024adaptive, liu2024attention, tu2024attribute} attempt to reconstruct missing views through data-level completion, feature-space recovery, or consensus-driven imputation by leveraging cross-view dependencies. While potentially effective when reconstructions are accurate, these methods suffer from a critical vulnerability: error propagation. Poor imputations distort structural patterns, which in turn degrade subsequent imputations, creating a self-reinforcing cycle of deteriorating clustering performance. 

On the other hand, imputation-free approaches ~\cite{liu2024sample, dai2025imputation} avoid explicit data recovery by learning shared latent spaces directly from available views or performing clustering without view reconstruction. These methods demonstrate greater robustness under moderate missingness but face escalating challenges as missing data increases: (i) instance misalignment, where missing views disrupt sample correspondences, hindering the construction of a coherent latent space \cite{ziyu2025deep}; (ii) information imbalance, where dominant views overshadow less-represented ones, distorting feature learning and diminishing the contribution of incomplete views \cite{xu2024deep}; and (iii) uncertainty in representation learning and clustering assignments, where high missingness weakens confidence in sample-to-cluster assignments \cite{chen2025biased}. This uncertainty is particularly critical for instances positioned near cluster boundaries or within closely related but distinct clusters, where small variations in representation can lead to misclassification during clustering. The resulting unreliable clustering decisions exacerbate inaccuracies in both learned features and the identification of latent patterns.

To address these challenges, we propose Deep Incomplete Multi-View Clustering via Hierarchical Imputation and Alignment (DIMVC-HIA), which integrates view-specific representation learning with a two-stage hierarchical imputation strategy and dual alignment mechanisms. As illustrated in Figure 1, our framework first employs view-specific autoencoders to learn latent representations coupled with a shared predictor for soft clustering assignments, then performs hierarchical imputation by (i) recovering missing cluster assignment distributions through cross-view contrastive similarity to establish semantic priors, followed by (ii) structure-preserving latent feature reconstruction using intra-view, intra-cluster statistics. To ensure robust clustering, we introduce dual alignment: energy-based models (EBMs) \cite{bachtis2024cascade,peng2024novel} minimize intra-cluster energy variance to enhance semantic coherence, while a contrastive objective aligns cluster assignments across views to maintain inter-view consistency and promote well-separated clusters. This unified architecture addresses the core challenges of error propagation in imputation-based methods and representation uncertainty in imputation-free approaches, achieving both accurate data recovery and reliable clustering. Our contributions can be summarized as follows:
\begin{itemize}
    \item We propose DIMVC-HIA, a novel DIMVC framework that jointly optimizes view-specific representation learning, hierarchical imputation, and dual-alignment regularization within an end-to-end architecture, effectively bridging the gap between imputation-based and imputation-free paradigms.
    \item We develop a hierarchical imputation strategy that first recovers missing cluster assignments using inter-view contrastive similarity, and then reconstructs missing features conditioned on these imputed cluster structures using intra-view cluster statistics, ensuring both geometric and semantic consistency.
    \item We design dual alignment modules: (i) an energy-based alignment that promotes compact clusters using cluster-specific energy models, and (ii) a contrastive assignment alignment that improves inter-view consistency and encourages confident, well-separated predictions. 
\end{itemize}

\begin{figure*}[ht!]
  \centering
  \includegraphics[width=0.92\textwidth]{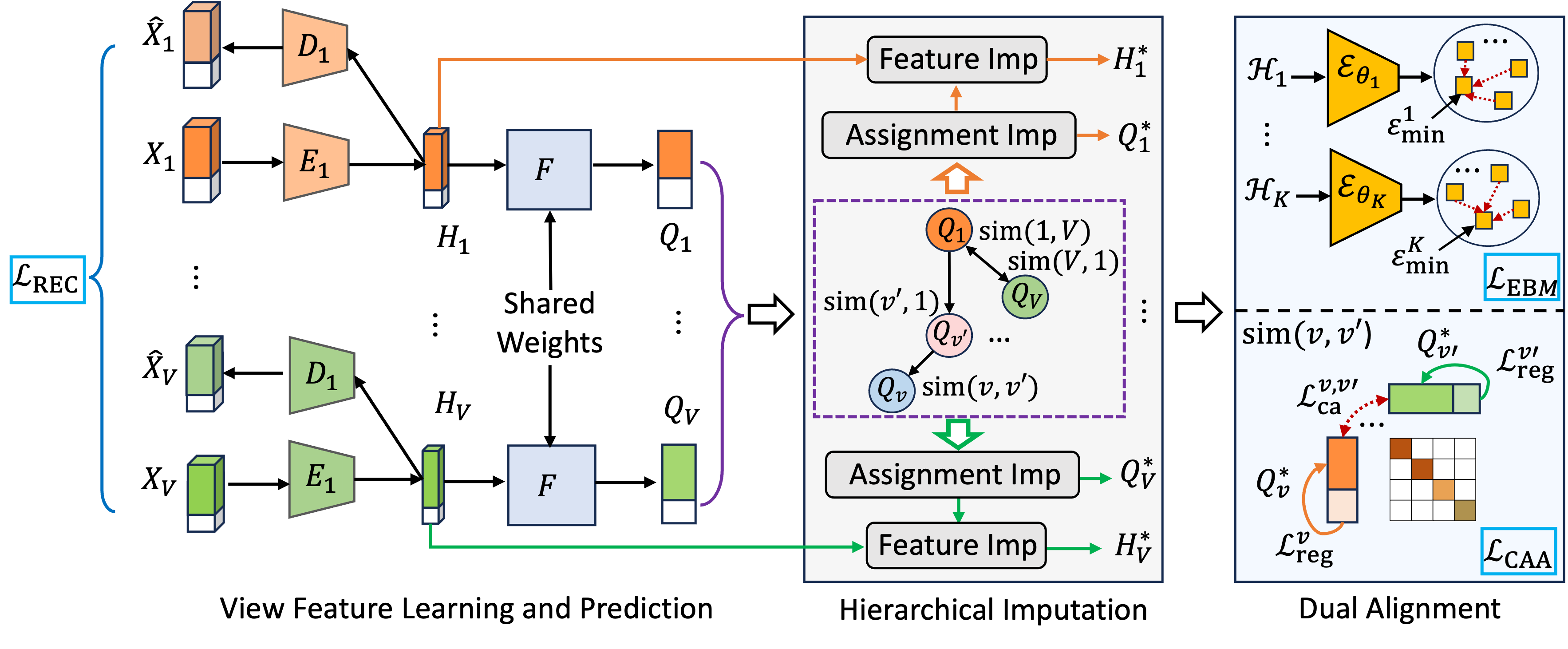}
  \caption{The overall architecture of our proposed DIMVC-HIA framework.}
  \label{fig:image}
\end{figure*}

\section{Related Work}

\subsection{Deep Incomplete Multi-View Clustering}
DIMVC leverages the powerful representation learning capabilities of deep neural networks such as autoencoders (AEs), generative adversarial networks (GANs), and graph neural networks (GNNs) to reduce dimensionality and learn shared latent structures across views, enabling effective handling of incomplete or missing data. Among various strategies, contrastive learning~\cite{shiri2024supervised, sun2024motif} has emerged as a prominent and widely adopted self-supervised learning approach. Its fundamental objective is to maximize the agreement between positive pairs while minimizing the similarity between negative pairs \cite{dou2025local}. This can be appl11ied at the instance level, cluster level, or both, to enhance feature representation and improve clustering performance. For instance, Wang et al. \cite{wang2025deep} employed cross-view contrastive alignments at both the instance and cluster levels, effectively capturing consistent assignments. Xue et al. \cite{xue2022robust} proposed a diversified graph contrastive regularization strategy at the intra-graph, inter-graph, and clustering levels to capture diverse data correlations, enhance discriminative representation learning, and mitigate information loss.

In addition, many DIMVC methods adopt imputation-based strategies~\cite{yang2025graph, kim2025predict} to recover missing views and enable more complete representation learning. Some approaches infer missing data by exploiting structural relationships across views, such as sample similarity, graph connectivity, or prototype semantics, often using tools like GNNs or clustering-based alignment~\cite{li2021contrastive, zhang2023unified}. Others adopt predictive or generative models to reconstruct missing features from observed ones~\cite{li2019gan}. Representative methods such as MICA~\cite{ziyu2025deep} follow this paradigm by inferring assignments from imputed features, making it directly affected by imputation noise. While these methods have shown promise, they face challenges when observed views are noisy or insufficient, potentially leading to inaccurate imputation or distribution mismatch.

\subsection{Energy-based Models}

EBMs offer a principled approach to modeling complex data distributions by associating each data sample with a scalar energy value that reflects its plausibility \cite{lecun2006tutorial, salakhutdinov2009deep}. Formally, a probability density $p_\theta(x)$ for $x \in \mathbb{R}^d$ can be defined via the Boltzmann distribution:
\begin{equation}\label{eq:ebm}
p_\theta(x) = \frac{\exp(-\mathcal{E}_\theta(x))}{Z_\theta},
\end{equation}
where $\mathcal{E}_\theta(x): \mathbb{R}^d \rightarrow \mathbb{R}$ denotes the parameterized energy function that assigns a scalar energy to each input $x$, and $Z_\theta = \int \exp(-\mathcal{E}_\theta(x))dx$ is the partition function that ensures proper normalization. In practice, computing $Z_\theta$ is intractable for high-dimensional data, but this is often unnecessary. Many training and inference procedures avoid direct dependence on $Z_\theta$ by using energy differences or relative probabilities, which cancel out the partition function~\cite{nie2021controllable, margeloiu2024tabebm}. This allows EBMs to directly shape the energy landscape--assigning lower energy to likely samples and higher energy to unlikely ones--without requiring explicit normalization.

\section{Methodology}
\label{sec:methodology}

\textbf{Notations.}
Let $\text{X} = \{X_v \in \mathbb{R}^{N \times d_v}\}_{v=1}^{V}$ denote an incomplete multi-view dataset with $V$ views and up to $N$ samples per view. For each view $v$, $X_v$ is the data matrix, where the $i$-th sample is $X_v(i) \in \mathbb{R}^{d_v}$ and its $(i,j)$-th entry is $X_v(i,j)$. The available samples in view $v$ are denoted as $X_v^a \in \mathbb{R}^{N_v \times d_v}$ with $N_v \leq N$, and missing entries are padded with zeros. A binary indicator matrix $G \in \{0, 1\}^{N \times V}$ marks sample availability, with $G(i,v) = 1$ if $X_v(i)$ is observed, and $0$ otherwise. The goal is to uncover $K$ shared cluster structures across views.

\subsection{View Feature Learning and Prediction}
We first adopt view-specific autoencoders to extract features from heterogeneous multi-view data. For data $X_v$ from view $v$, an encoder $E_v$ and decoder $D_v$ are used to extract latent features $H_v \in \mathbb{R}^{N \times d}$, and reconstruct the input: 
\begin{equation}\label{eq:1.1}
H_v = E_v(X_v; \phi_v^e), \quad \widehat{X}_v = D_v(H_v; \phi_v^d). 
\end{equation}
Here, $d$ is the dimension of the shared latent space, and $\theta_v$, $\phi_v$ are the parameters of the encoder and decoder, respectively. Each AE is pre-trained independently by minimizing the reconstruction loss over available samples:
\begin{equation}\label{eq:REC}
\mathcal{L}_{\text{REC}} = \frac{1}{VN}\sum_{v=1}^V \sum_{i=1}^{N} G(i,v) \left\| X_v(i) - \widehat{X}_v(i) \right\|_2^2.
\end{equation}
The indicator $G(i,v)$ ensures that only observed samples contribute to the loss. 

The latent features $H_v$ are subsequently mapped to soft cluster assignments via a shared clustering predictor $F$ parameterized by $\vartheta$:
\begin{equation}\label{eq:view_soft_labels}
    Q_v(i) =F({H}_v(i); \vartheta)
    \in
    \mathbb{R}^{K}, 
\end{equation}
where $Q_v(i)$ denotes the cluster assignment distribution for the $i$-th sample in view $v$.

\subsection{Hierarchical Imputation}
We argue that the latent space of soft cluster assignments encodes richer semantic information that more directly reflects the underlying cluster structures. 
Therefore, we propose to first explore cross-view relationships within this assignment space and leverage them to guide the imputation of missing cluster assignments. The imputed assignments are then used to inform the imputation of latent feature representations.

\subsubsection{Cluster Assignment Imputation.}
To impute missing cluster assignments, we exploit semantic relationships across views by constructing pairwise similarity graphs from co-observed samples. The key idea is that samples observed in multiple views should share consistent clustering semantics. These correspondences enable us to infer missing assignments in one view using reliable predictions from semantically aligned views. For each distinct view pair $(v, v')$, where $v \ne v'$ and $v, v' \in \{1, \cdots, V\}$, we define the index set of co-observed samples as:
\begin{equation}
\mathcal{I}_{v,v'} = \left\{\, i \;\middle|\; G(i,v)=1 \text{ and } G(i,v')=1 \,\right\}
\end{equation}
Let $|\mathcal{I}_{v,v'}|$ be the number of such paired samples.  We then extract the soft cluster assignments for these samples in views $v$ and $v'$ as:
\begin{equation}
    Q_v^{v,v'} = [Q_v(j)]_{j \in \mathcal{I}_{v,v'}}, \;
    Q_{v'}^{v',v} = [Q_{v'}(j)]_{j \in \mathcal{I}_{v,v'}}
\end{equation}
where $Q_v^{v,v'}, Q_{v'}^{v',v} \in \mathbb{R}^{|\mathcal{I}_{v,v'}| \times K}$. The cross-view similarity matrix between these two sets of assignments is computed as:
\begin{equation}
   S_{v,v'} = Q_v^{v,v'} (Q_{v'}^{v',v})^\top  \in
    \mathbb{R}^{|\mathcal{I}_{v,v'}| \times |\mathcal{I}_{v,v'}|}.
\end{equation}

To quantify the semantic alignment between two views, we propose a label-aware contrastive similarity score.
For each co-observed sample $i \in \mathcal{I}_{v,v'}$ in view $v$, the positive similarity is defined as the diagonal entry $S_{v,v'}(i, i)$, which measures agreement between the two views for the same instance. The negative similarities are given by off-diagonal entries $S_{v,v'}(i, j)$ where $j \ne i$ and the predicted cluster labels differ, i.e., $\hat{y}_v(i) \ne \hat{y}_{v'}(j)$. Here, $\hat{y}_v(i) = \arg\max_k Q_v(i,k)$ and $\hat{y}_{v'}(j) = \arg\max_k Q_{v'}(j,k)$. 

To avoid penalizing semantically consistent samples, we exclude false negatives--samples $j \neq i$ that share the same predicted label as $i$. 
Let $\mathcal{F}_{v,v'}^i =\{ j \in \mathcal{I}_{v,v'} \mid j \ne i, \hat{y}_v(i) = \hat{y}_{v'}(j)$ denote the set of false negatives in view $v'$ with respect to sample $i$ in view $v$. The set of valid negative candidates is then given by: $\mathcal{B}_{v,v'}^i = \{\mathcal{I}_{v,v'} \setminus \mathcal{F}_{v,v'}^i\}$.
Using these filtered negatives, the contrastive similarity score between views $v$ and $v'$ is computed as:
\begin{equation}\label{eq:sim}
\text{sim}{(v,v')} = \frac{1}{|\mathcal{I}_{v,v'}|} \sum_{i \in \mathcal{I}_{v,v'}}
\frac{\exp\left( S_{v,v'}(i,i)/\tau \right)}{\sum_{j \in \mathcal{B}_{v,v'}^i} \exp\left( S_{v,v'}(i,j)/\tau \right)}.
\end{equation}
This label-aware contrastive formulation improves semantic discriminability by highlighting inconsistencies and reducing the impact of noisy similarities. The resulting scores guide assignment imputation by prioritizing more semantically aligned views.

With the computed cross-view similarity scores $\text{sim}(\cdot,\cdot)$, each target view $v$ defines a reference view list $\mathcal{R}_v$, which contains all other views sorted in descending order of semantic similarity:
\begin{equation}
\mathcal{R}_v = \operatorname{argsort}_{v' \ne v} \left( \text{sim}{(v,v')} \right).
\end{equation}
For each missing sample $i$ in view $v$, we sequentially traverse the ranked list $\mathcal{R}_v$ and select the first view $v' \in \mathcal{R}_v$ such that sample $i$ is available in view $v'$. We denote this selected reference view as:
\begin{equation}
\pi_v^i =  \min \left\{ v' \in \mathcal{R}_v \mid G(i, v') = 1 \right\}.
\end{equation}
The imputed cluster assignment for sample $i$ in view $v$ is then obtained from the most semantically aligned available view $\pi_v^i$, and the completed assignment distribution is defined as:
\begin{equation}\label{eq:Q_complete}
Q^*_v(i) = 
\begin{cases}
Q_v(i), & \text{if } G(i, v) = 1 \\
Q_{\pi_v^i}(i), & \text{otherwise.}
\end{cases}
\end{equation}
This similarity-guided imputation strategy ensures that the missing cluster assignment probabilities are inferred from the semantically aligned and structurally reliable views. 

\subsubsection{Latent Feature Imputation.}
After completing the cluster assignments $Q^*_v$ for each view $v$ via inter-view similarity, we further leverage this information to impute missing latent features in a cluster-aware intra-view manner.

For each missing sample $i$ in view $v$, we first determine its most likely cluster label:
\begin{equation}
\hat{y}_v(i) = \arg\max_k Q^*_v(i, k), \; k \in \{1, \ldots, K\}. 
\end{equation}
We then compute the cluster prototype $\mathcal{C}_v(k)$ as the average latent feature of all available samples in view $v$ that are predicted to belong to cluster $k$:
\begin{equation}
\mathcal{C}_v(k) = \frac{1}{|Y_v^k|} \sum_{j \in Y_v^k} H_v(j)
\end{equation}
where $Y_v^k = \{j \mid G(j, v) = 1, \,\hat{y}_v(j)=k\}$. 
The completed latent feature $H^*_v(i)$ is then obtained by:
\begin{equation} \label{eq:H_complete}
H^*_v(i) =
\begin{cases}
H_v(i), & \text{if } G(i, v) = 1 \\
\mathcal{C}_v({\hat{y}_v(i)}), & \text{otherwise.} 
\end{cases}
\end{equation}
Although non-parametric, these imputations influence learning through dual-alignment losses, while reconstruction on observed samples stabilizes training.

\subsection{Energy-Based Semantic Alignment}

Given the completed features $H^*_v$ and predicted cluster labels $\hat{y}_v$, we introduce an energy-based semantic alignment module to enhance intra-cluster compactness and semantic consistency across views. This module leverages cluster-aware EBMs to assess the reliability of latent features--assigning lower energy to those more semantically aligned with their clusters.
For each cluster $k \in \{1,\cdots,K\}$, we define a view-shared, cluster-specific energy function $\mathcal{E}_{\theta_k}: \mathbb{R}^d \mapsto \mathbb{R}^+$, which  maps feature vectors to scalar energy scores. A lower score indicates stronger compatibility with cluster $k$. We first construct the cluster-level feature set by collecting all features across views that are assigned to cluster $k$:
\begin{equation}
\begin{aligned}
\mathcal{H}_k = \{\, H^*_v(i)\ \big|\ 
& \hat{y}_v(i) = k, \forall\, v, i \}. 
\end{aligned}
\end{equation}
To promote semantic alignment within each cluster, we minimize the energy deviation of features from the most confident cluster anchor--the feature with the lowest energy:
\begin{equation}
\varepsilon^k_{\text{min}} = \min_{\mathbf{h}' \in \mathcal{H}_k} \mathcal{E}_{\theta_k}(\mathbf{h}';\theta_k).
\end{equation}
The energy-based alignment loss for cluster $k$ is then computed to enforce intra-cluster compactness:
\begin{equation}
\mathcal{L}_{\text{EBM}}^k = \frac{1}{|\mathcal{H}_k|} \sum_{\mathbf{h} \in \mathcal{H}_k} \left| \mathcal{E}_{\theta_k}(\mathbf{h};\theta_k) - \varepsilon^k_{\text{min}} \right|. 
\end{equation}
The overall energy-based alignment loss is obtained by averaging over all clusters:
\begin{equation}\label{eq:EBM}
\mathcal{L}_{\text{EBM}} = \frac{1}{K} \sum_{k=1}^{K} \mathcal{L}_{\text{EBM}}^k.
\end{equation}
This formulation encourages features within each cluster to concentrate around reliable, low-energy anchors. It enables the model to flexibly shape continuous energy landscapes beyond center-based regularizers.

\subsection{Contrastive Assignment Alignment}
Given the completed soft cluster assignments $Q^*_v$, we introduce a contrastive assignment alignment (CAA) loss that promotes both sample-level semantic agreement and confident, well-separated cluster predictions. The total loss is defined as:
\begin{equation}\label{Eq:CAA}
\mathcal{L}_{\text{CAA}} = \frac{1}{2} \sum_{v=1}^V \sum_{v'=1, v' \ne v}^V [\text{sim}(v,v')\cdot \mathcal{L}_{\text{ca}}^{v,v'} + \mathcal{L}_{\text{reg}}^{v,v'}].
\end{equation}
Here, $\text{sim}(v,v')$ represents the semantic similarity score between views $v$ and $v'$, as defined in Eq. (\ref{eq:sim}). The alignment loss comprises two components. 

The contrastive alignment loss is used to pull together the cluster assignment distributions of matching samples across views:
\begin{equation}
\mathcal{L}_{\text{ca}}^{v,v'} = -\frac{1}{N}\sum_{i=1}^N\log \frac{\exp(Q^*_v(i)^\top Q^*_{v'}(i)/\tau)}{\sum_{j=1}^N\exp(Q^*_v(i)^\top Q^*_{v'}(j)/\tau) },
\end{equation}
where $\tau$ is a temperature scaling factor. To promote confident and balanced predictions, we incorporate a distribution-level entropy regularization. For each view, we compute the average cluster assignment distribution: 
\begin{equation}
q_v(k) = \frac{1}{N}\sum_{i=1}^N Q^*_v(i,k), \; q_{v'}(k) = \frac{1}{N}\sum_{j=1}^N Q^*_{v'}(j,k). 
\end{equation}
The entropy regularization term is then given by:
\begin{equation}\label{eq:distribution_entropy_loss}
\mathcal{L}_{\text{reg}}^{v,v'} = \frac{1}{K}\sum_{k=1}^{K} [q_v(k) \log\left( q_v(k) \right) + q_{v'}(k) \log\left( q_{v'}(k) \right)].
\end{equation}
The entropy regularizer promotes balanced cluster proportions and prevents trivial all-in-one or empty assignments.

\subsection{Overall Objective}
The overall training objective function is defined as:
\begin{equation}\label{eq:total}
\mathcal{L} = \mathcal{L}_{\text{REC}} + \alpha \cdot \mathcal{L}_{\text{EBM}} + \beta \cdot  \mathcal{L}_{\text{CAA}}.
\end{equation}
Here, $\mathcal{L}_{\text{REC}}$ denotes the reconstruction loss for pretraining the autoencoders, $\mathcal{L}_{\text{EBM}}$ encourages intra-cluster compactness across views, and $\mathcal{L}_{\text{CAA}}$ enforces cross-view consistency and confidence through contrastive assignment alignment. The hyperparameters $\alpha$ and $\beta$ balance the contribution of the alignment terms. The full set of trainable model parameters, denoted as $\Phi$, includes the view-specific autoencoder parameters $\{\phi_v^e, \phi_v^d\}_{v=1}^V$, the view-shared clustering predictor parameters $\vartheta$, and the cluster-specific EBM parameters $\{\theta_k\}_{k=1}^K$. The training procedure is summarized in Algorithm 1. After training, the final cluster label for sample $i$ is determined by $y_i=\arg\max_k(\sum_v Q^*_v(i,k))$, and the final labels $Y=[y_1,\cdots,y_N]$. 

\begin{algorithm}[t]
\caption{Training Procedure of DIMVC-HIA}
\label{alg:method}
\begin{algorithmic}[1]
  \STATE \textbf{Input:} Incomplete multi-view data $\text{X}$, missing indicator matrix $G$, hyperparameters $\alpha, \beta, \tau$
  \STATE \textbf{Output:} Final clustering assignments $Y$
  \STATE Initialize model parameters $\Phi$ 
  \STATE Pretrain autoencoders by minimizing $\mathcal{L}_{\text{REC}}$ via Eq. (\ref{eq:REC})
  \FOR{$t=1$ to $T$}
    \STATE Sample a mini-batch $\{X_v(i)\}_{i=1}^B$ from each view $X_v$
    \STATE Compute features $\{H_v(i)\}_{i=1}^B$ and soft cluster assignments $\{Q_v(i)\}_{i=1}^B$ via Eqs. (\ref{eq:1.1}) and (\ref{eq:view_soft_labels}), respectively
    \STATE Compute inter-view semantic similarity scores $\text{sim}(\cdot,\cdot)$ via Eq. (\ref{eq:sim})
    \STATE Generate completed cluster assignments $\{Q^*_v(i)\}_{i=1}^B$ for each view via Eq. (\ref{eq:Q_complete})
    \STATE Generate completed features $\{H^*_v(i)\}_{i=1}^B$ for each view via Eq. (\ref{eq:H_complete})
    \STATE Compute $\mathcal{L}_{\text{EBM}}$ and $\mathcal{L}_{\text{CAA}}$ via Eqs. (\ref{eq:EBM}) and (\ref{Eq:CAA})
    \STATE Compute the total training loss $\mathcal{L}$ via Eq. (\ref{eq:total})
    \STATE Update model parameters $\Phi$ using backpropagation
  \ENDFOR
  \STATE Compute final cluster labels $Y$ 
\end{algorithmic}
\end{algorithm}

\section{Experiments}

\subsection{Experimental settings}

\subsubsection{Datasets and Evaluation Metrics}

We conducted experiments on four benchmark multi-view datasets: BDGP~\cite{cai2012joint}, MNIST-USPS~\cite{asuncion2007uci}, Fashion-MNIST~\cite{xiao2017fashion}, and Handwritten~\cite{asuncion2007uci}, as summarized in Table~\ref{tab:datasets}. The incomplete versions of these datasets are generated by randomly omitting a proportion of samples from each view, corresponding to missing ratios $\eta \in \{0.1, 0.3, 0.5, 0.7\}$. Three metrics, ACC, NMI, and Purity (PUR), are adopted to evaluate the clustering performance.

\begin{table}[h]
\centering
\small
\begin{tabular}{lcl}
\toprule
\textbf{Dataset} & \textbf{N / V / K} & \textbf{Dimensions} \\
\midrule
BDGP         & 2,500 / 2  / 5   & 1,750/79            \\
MNIST–USPS   & 5,000 / 2  / 10   & 784/256             \\
Fashion-MNIST  & 10,000 / 3 / 10  & 784/784/784         \\
Handwritten  & 2,000 / 6 / 10    & 240/76/216/47/64/6  \\
\bottomrule
\end{tabular} \\
{\footnotesize \textbf{N} (\# of Samples); \textbf{V} (\# of Views); \textbf{K} (\# of Clusters)}.
\caption{Statistics of the benchmark datasets.}
\label{tab:datasets}
\end{table}

\begin{table*}[ht!]
\centering
\footnotesize
\begin{tabular}{|c|l|ccc|ccc|ccc|ccc|}
\toprule
\multirow{2}{*}{\textbf{Dataset}} & \multirow{2}{*}{\textbf{Method}} & \multicolumn{3}{c|}{$\eta = 0.1$} & \multicolumn{3}{c|}{$\eta = 0.3$} & \multicolumn{3}{c|}{$\eta = 0.5$} & \multicolumn{3}{c|}{$\eta = 0.7$} \\
\cmidrule{3-14}
& & ACC & NMI & PUR & ACC & NMI & PUR & ACC & NMI & PUR & ACC & NMI & PUR \\
\midrule
\multirow{11}{*}{\rotatebox[origin=c]{90}{\textbf{BDGP}}} 
& COMPLETER & 89.88 & 64.81 & 89.88 & 64.60 & 42.66 & 65.72 & 57.32 & 34.63 & 58.80 & 50.80 & 31.01 & 52.04 \\
& CPSPAN & 65.00 & 65.10 & 71.52 & 75.92 & 65.82 & 75.92 & 63.64 & 54.27 & 69.68 & 72.48 & 63.07 & 72.48 \\
& DCG & 95.38 & 87.00 & 95.38 & 91.59 & 76.89 & 91.59 & 76.47 & 58.03 & 76.47 & 66.00 & 47.23 & 66.00 \\
& DIVIDE & 89.68 & 64.61 & 89.68 & 64.40 & 42.46 & 65.52 & 57.12 & 34.43 & 58.60 & 51.60 & 30.81 & 51.84 \\
& RPCIC & 88.08 & 63.01 & 88.08 & 62.80 & 40.86 & 63.92 & 55.52 & 32.83 & 57.00 & 50.00 & 29.21 & 50.24 \\
& APADC & 65.52 & 55.89 & 65.52 & 82.48 & 68.28 & 82.48 & 67.16 & 56.14 & 67.60 & 64.36 & 51.10 & 64.36 \\
& GIMVC & 72.21 & 60.95 & 73.74 & 72.10 & 57.85 & 72.98 & 69.92 & 57.28 & 72.04 & 63.48 & 50.93 & 65.44 \\
& PMIMC & 87.96 & 72.14 & 87.96 & 89.52 & 75.55 & 89.52 & 93.04 & 80.92 & 93.04 & \underline{91.72} & 77.78 & \underline{91.72} \\
& ProImp & 75.20 & 70.21 & 77.12 & 92.56 & 82.20 & 92.56 & 90.92 & 77.76 & 90.92 & 84.72 & 65.63 & 84.72 \\
& DSIMVC & \underline{98.00} & \underline{93.50} & \underline{98.00} & \underline{96.08} & \underline{88.56} & \underline{96.08} & \underline{93.56} & \underline{84.04} & \underline{93.56} & 91.12 & \underline{79.00} & 91.12 \\
& DIMVC-HIA & \textbf{98.40} & \textbf{94.49} & \textbf{98.40} & \textbf{96.25} & \textbf{89.97} & \textbf{96.25} & \textbf{95.16} & \textbf{85.14} & \textbf{95.16} & \textbf{92.32} & \textbf{79.50} & \textbf{92.32} \\

\midrule

\multirow{11}{*}{\rotatebox[origin=c]{90}{\textbf{MNIST-USPS}}} 
& COMPLETER & 78.20 & 81.37 & 81.76 & 71.06 & 68.74 & 71.34 & 64.20 & 58.89 & 64.38 & 51.08 & 46.19 & 51.66 \\
& CPSPAN & 81.52 & 80.45 & 84.58 & 65.74 & 72.42 & 74.96 & 74.36 & 78.40 & 79.32 & 85.16 & 80.68 & 85.18 \\
& DCG & \underline{99.05} & 97.13 & \underline{99.05} & \underline{97.48} & 93.21 & \underline{97.48} & 96.09 & 90.11 & 96.09 & 92.58 & 83.17 & 92.58 \\
& DIVIDE & 75.80 & 78.57 & 79.46 & 73.03 & 69.54 & 73.31 & 63.90 & 58.59 & 64.08 & 50.78 & 45.89 & 51.36 \\
& RPCIC & 77.05 & 80.37 & 81.06 & 69.36 & 67.04 & 69.64 & 62.50 & 57.19 & 62.68 & 49.38 & 44.49 & 49.96 \\
& APADC & 97.32 & 93.58 & 97.32 & 96.24 & 90.87 & 96.24 & 94.16 & 86.61 & 94.16 & 91.32 & 82.51 & 91.32 \\
& GIMVC & 77.63 & 74.37 & 79.02 & 76.05 & 70.93 & 77.19 & 74.55 & 66.03 & 75.13 & 74.18 & 60.73 & 74.18 \\
& PMIMC & 77.34 & 76.27 & 79.74 & 72.48 & 72.15 & 74.76 & 76.14 & 75.63 & 79.88 & 78.94 & 74.40 & 79.12 \\
& ProImp & 99.02 & \underline{97.56} & 99.02 & 97.46 & 93.82 & 97.46 & 96.32 & \underline{91.10} & 96.32 & 93.42 & \textbf{86.17} & 93.42 \\
& DSIMVC & 98.00 & 94.98 & 98.00 & 97.08 & \textbf{93.92} & 97.08 & \underline{96.44} & 91.01 & \underline{96.44} & \underline{92.76} & \underline{85.29} & \underline{92.76} \\
& DIMVC-HIA & \textbf{99.10} & \textbf{97.61} & \textbf{99.10} & \textbf{97.54} & \underline{93.84} & \textbf{97.54} & \textbf{96.48} & \textbf{91.20} & \textbf{96.48} & \textbf{93.66} & 82.16 & \textbf{93.66} \\

\midrule

\multirow{11}{*}{\rotatebox[origin=c]{90}{\textbf{Fashion-MNIST}}} 
& COMPLETER & 70.55 & 84.92 & 78.91 & 64.92 & 68.27 & 68.09 & 60.64 & 61.36 & 61.80 & 55.41 & 49.24 & 49.31 \\
& CPSPAN & 66.63 & 72.51 & 71.97 & 76.64 & 77.66 & 80.80 & 60.82 & 70.44 & 67.44 & 59.75 & 69.68 & 66.00 \\
& DCG & 95.96 & 87.87 & 95.96 & 92.68 & 82.55 & 92.68 & 90.15 & 82.88 & 90.15 & 84.46 & 76.10 & 84.46 \\
& DIVIDE & 70.45 & 84.82 & 78.81 & 64.82 & 68.17 & 67.99 & 60.54 & 61.26 & 61.70 & 55.31 & 49.14 & 49.21 \\
& RPCIC & 68.85 & 83.22 & 77.21 & 63.22 & 66.57 & 66.39 & 58.94 & 59.66 & 60.10 & 43.71 & 47.54 & 47.61 \\
& APADC & 66.10 & 77.08 & 67.80 & 60.58 & 73.38 & 65.41 & 57.08 & 69.23 & 57.89 & 64.63 & 67.77 & 66.83 \\
& GIMVC & 75.98 & 77.87 & 78.98 & 75.47 & 74.46 & 78.36 & 69.65 & 69.09 & 73.63 & 67.46 & 64.86 & 71.37 \\
& PMIMC & 71.03 & 77.25 & 75.66 & 71.81 & 76.39 & 75.23 & 70.22 & 76.01 & 74.88 & 73.57 & 75.38 & 75.66 \\
& ProImp & \underline{96.26} & \underline{92.45} & \underline{96.26} & \underline{93.48} & \underline{87.89} & \underline{93.48} & \underline{91.01} & \underline{83.99} & \underline{91.01} & \underline{86.74} & \underline{78.19} & \underline{86.74} \\
& DSIMVC & 81.88 & 81.77 & 81.88 & 82.39 & 79.62 & 82.39 & 74.76 & 75.83 & 74.76 & 81.15 & 75.93 & 81.15 \\
& DIMVC-HIA & \textbf{98.84} & \textbf{97.12} & \textbf{98.84} & \textbf{97.16} & \textbf{93.72} & \textbf{97.16} & \textbf{96.51} & \textbf{92.25} & \textbf{96.51} & \textbf{95.27} & \textbf{92.05} & \textbf{95.27} \\

\midrule

\multirow{11}{*}{\rotatebox[origin=c]{90}{\textbf{HandWritten}}} 
& COMPLETER & 86.10 & 85.59 & 87.40 & 76.15 & 75.52 & 76.30 & 67.60 & 63.99 & 68.05 & 61.70 & 57.49 & 62.10 \\
& CPSPAN & 49.95 & 58.28 & 55.65 & 50.50 & 57.62 & 55.15 & 42.70 & 52.47 & 48.95 & 43.35 & 52.00 & 48.60 \\
& DCG & 86.74 & 84.30 & 86.74 & 82.07 & 77.28 & 82.08 & 81.13 & 76.71 & 81.13 & 79.08 & 72.45 & 79.08 \\
& DIVIDE & 85.90 & 85.39 & 87.20 & 75.95 & 75.32 & 76.10 & 67.40 & 63.79 & 67.85 & 61.50 & 57.29 & 61.90 \\
& RPCIC & 84.30 & 83.79 & 85.60 & 74.35 & 73.72 & 74.50 & 65.80 & 62.19 & 66.25 & 59.90 & 55.69 & 60.30 \\
& APADC & 80.75 & 83.02 & 82.95 & 82.35 & 83.29 & 83.95 & 83.35 & \underline{84.35} & 83.90 & 78.05 & 75.49 & 79.75 \\
& GIMVC & \underline{92.14} & \underline{88.34} & \underline{93.02} & \underline{93.58} & \underline{87.03} & \underline{93.58} & \underline{90.73} & 84.03 & \underline{91.21} & \underline{86.10} & \underline{80.41} & \underline{87.50} \\
& PMIMC & 81.55 & 78.88 & 81.80 & 91.75 & 85.21 & 91.75 & 78.65 & 78.79 & 82.20 & 77.65 & 78.26 & 80.70 \\
& ProImp & 82.25 & 78.79 & 82.25 & 80.10 & 76.80 & 80.10 & 81.00 & 77.56 & 81.00 & 78.90 & 73.29 & 78.90 \\
& DSIMVC & 75.55 & 71.30 & 79.15 & 73.10 & 70.53 & 75.70 & 71.00 & 67.87 & 75.70 & 65.35 & 64.44 & 69.00 \\
& DIMVC-HIA & \textbf{96.85} & \textbf{92.82} & \textbf{96.85} & \textbf{96.35} & \textbf{91.87} & \textbf{96.35} & \textbf{95.15} & \textbf{90.41} & \textbf{95.15} & \textbf{94.05} & \textbf{88.33} & \textbf{94.05} \\

\bottomrule
\end{tabular}
\caption{Performance comparison with state-of-the-art methods across different datasets with varying missing rates $\eta$. The best results are bold, and the second-best results are underlined.}
\label{tab:main_results}
\end{table*}

\subsubsection{Baseline Methods and Evaluation Metrics}
We compare the performance of DIVMC-HIA against ten state-of-the-art incomplete multi-view clustering methods. The evaluated baselines include: COMPLETER~\cite{lin2021completer}, CPSPAN~\cite{jin2023deep}, DCG~\cite{zhang2025incomplete}, DIVIDE~\cite{lu2024decoupled}, 
RPCIC~\cite{yuan2024robust}, APADC~\cite{xu2023adaptive}, GIMVC~\cite{bai2024graph}, PMIMC~\cite{yuan2025prototype}, ProImp~\cite{li2023incomplete}, and DSIMVC~\cite{tang2022deep}.

\subsubsection{Implementation Details}

DIMVC-HIA employs view-specific encoders with architectures \([d_v, 256, 512, d]\) for BDGP and Fashion-MNIST, and \([d_v, 256, 512, 1024, d]\) for MNIST-USPS and Handwritten, where \(d_v\) denotes the input dimension and \(d = 2000\) is fixed across all datasets. The shared MLP comprises two linear layers with dimensions \([d, 1024, K]\), where \(K\) is the number of clusters. Each cluster-specific EBM is an MLP with hidden dimensions \([256, 256, 256]\). We use batch sizes of 200 for BDGP and Handwritten, 100 for Fashion-MNIST, and 50 for MNIST-USPS, with a fixed learning rate of 0.0001. The training procedure includes 100 epochs of pre-training followed by 200 epochs of fine-tuning. Balancing coefficients are empirically set as $\alpha$ = 0.1 and $\beta$ = 0.01 for all datasets. The model is implemented using the PyTorch framework and trained on an NVIDIA RTX 3080 GPU with 32 GB RAM.

\subsection{Experimental Results and Analysis}
Table~\ref{tab:main_results} shows the performance of DIMVC-HIA across missing-view ratios $\eta \in [0.1, 0.7]$. The model outperforms most of the ten baselines, with its advantage becoming more pronounced as incompleteness increases. At $\eta = 0.7$, it achieves 95.27\% ACC on Fashion (8.53\% higher than ProImp) and 94.05\% on HandWritten (7.95\% higher than GIMVC). On BDGP and MNIST-USPS, it maintains the best overall performance with an accuracy variation of only around 6\%, demonstrating strong stability. In comparison, baseline methods such as DSIMVC experience significant drops in performance; for instance, its ACC on BDGP decreases from 98.00\% to 91.12\% as $\eta$ increases. These results confirm the effectiveness of DIMVC-HIA's hierarchical imputation in enabling robust and discriminative representation learning under severe view incompleteness.

\begin{figure}[ht]
  \centering
  \includegraphics[width=0.44\textwidth]{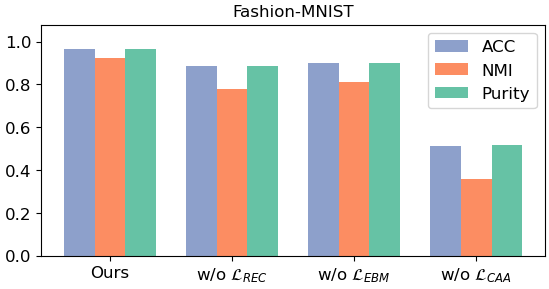}
  \caption{Ablation study results on the Fashion-MNIST dataset with $\eta$ = 0.5.}
  \label{fig:Ablation}
\end{figure}

\subsection{Ablation Study}
To further evaluate the effectiveness of the proposed DIMVC-HIA method, we performed an ablation study on the Fashion-MNIST dataset set with missing ratios ($\eta$) set to 0.5. We selectively removed key components including the reconstruction loss ($\mathcal{L}_{\text{REC}}$), the energy-based alignment loss ($\mathcal{L}_{\text{EBM}}$), and the contrastive assignment alignment loss ($\mathcal{L}_{\text{CAA}}$) in Eq. (\ref{eq:total}). The results are presented in Figure~\ref{fig:Ablation}. Notably, removing the contrastive assignment alignment loss ($\mathcal{L}_{\text{CAA}}$) causes the most significant performance drop, highlighting its critical role in aligning clustering assignments. Meanwhile, excluding either the energy-based alignment loss ($\mathcal{L}_{\text{EBM}}$) or the reconstruction loss ($\mathcal{L}_{\text{REC}}$) also results in considerable degradation, confirming their importance in ensuring stable optimization and learning robust, discriminative representations.

\subsection{Visualization Results and Convergence Analysis}
\begin{figure}[ht!]
  \centering
  \includegraphics[width=0.21\textwidth]{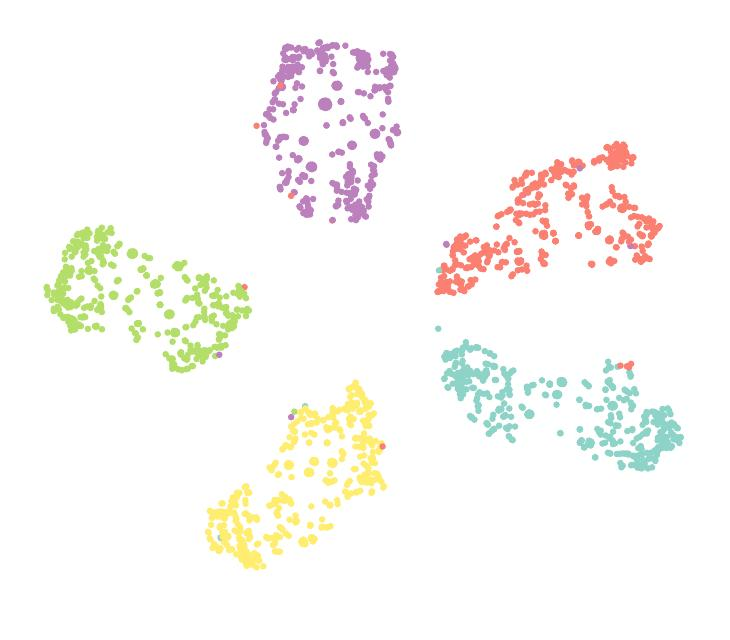}
  \includegraphics[width=0.21\textwidth]{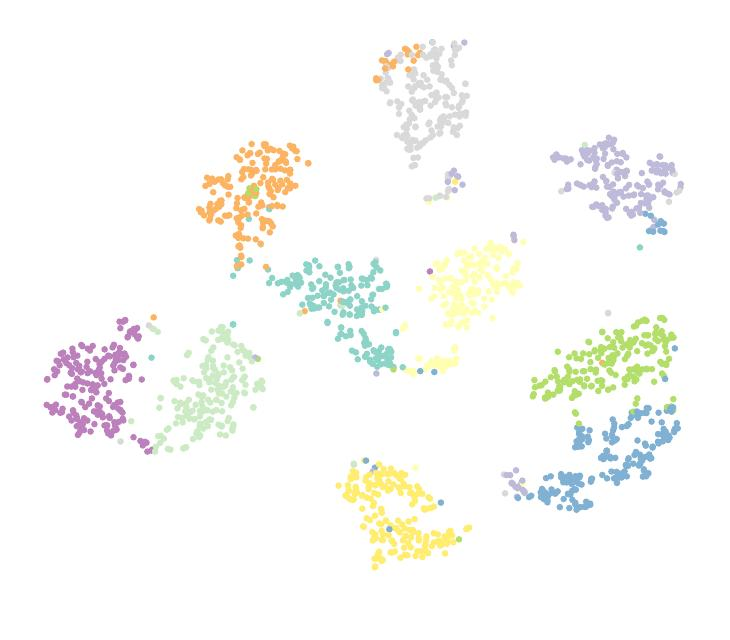}
  \caption{T-SNE visualizations of embeddings on BDGP with $\eta = 0.5$ (left) and Handwritten with $\eta = 0.7$ (right).}
  \label{fig:TSNE}
\end{figure}

To evaluate the clustering performance of DIMVC-HIA, Fig. \ref{fig:TSNE} presents the T-SNE visualization of the complete feature embeddings ($H^*$) on the BDGP and Handwritten datasets, with missing ratios of $\eta = 0.5$ and $\eta = 0.7$, respectively. The t-SNE visualization on the BDGP dataset reveals distinct and compact clusters with well-defined inter-class boundaries, aligning with its high clustering accuracy. The clear separation and minimal overlap between clusters reflect the model’s strong discriminative capability. Despite the high missing ratio and a larger number of clusters, the t-SNE result on the Handwritten dataset also exhibits visibly separable cluster structures. 

\begin{figure}[ht!]
  \centering
  \includegraphics[width=0.23\textwidth]{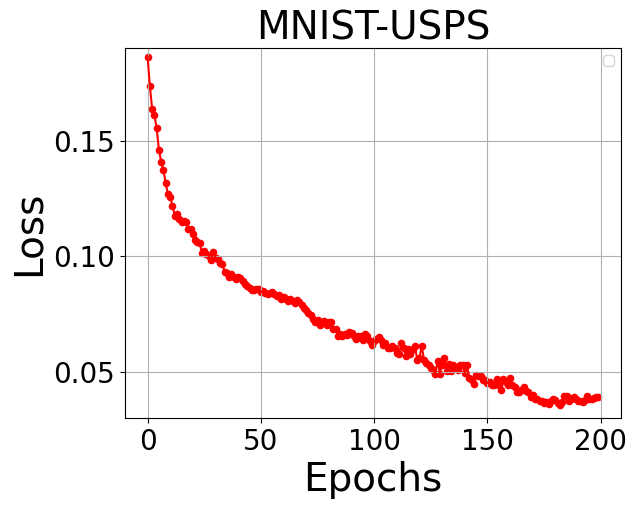}
  \includegraphics[width=0.23\textwidth]{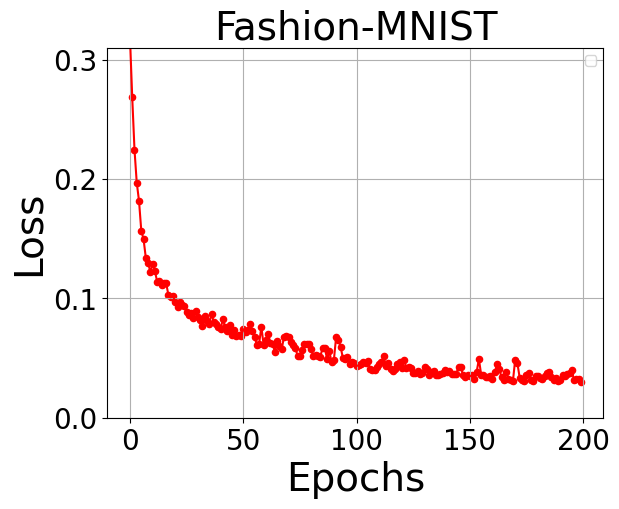}
  \caption{Convergence curves for the MNIST-USPS and Fashion-MNIST datasets with $\eta = 0.1$.}
  \label{fig:convergence}
\end{figure}

To evaluate the convergence behavior of DIMVC-HIA, Figure~\ref{fig:convergence} shows the loss curves on the MNIST-USPS and Fashion-MNIST datasets under $\eta = 0.1$. The loss exhibits a consistent downward trend throughout training, reflecting stable convergence. In both datasets, a rapid decline occurs within the first 25 epochs, followed by a slower, steady reduction that eventually converges to a stable value. This pattern demonstrates the effectiveness of DIMVC-HIA’s optimization strategy and highlights the robustness of its architecture in facilitating efficient and reliable training.

\begin{figure}[ht!]
  \centering
  \includegraphics[width=0.233\textwidth]{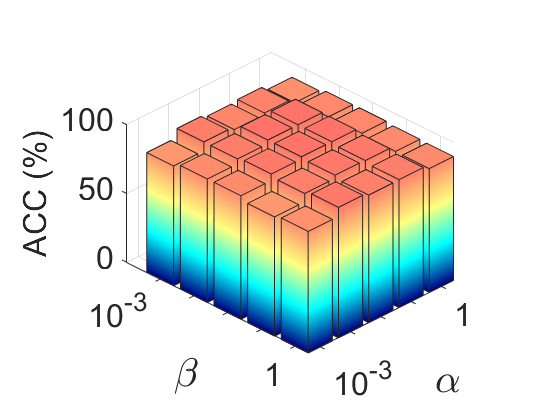}
  \includegraphics[width=0.233\textwidth]{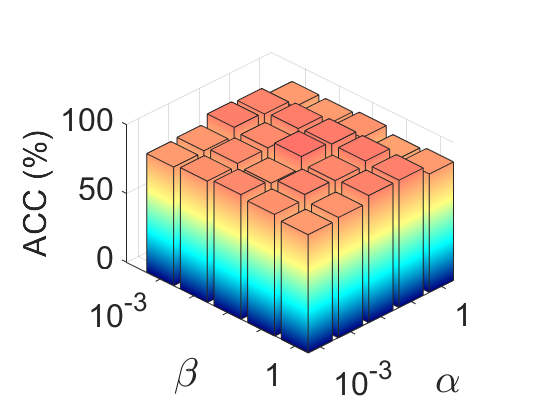}
  \caption{Parameter sensitivity of $\alpha$ and $\beta$ on BDGP (left) and Handwritten (right) datasets with $\eta = 0.3$, where both $\alpha$ and $\beta$ take values in $\{0.001, 0.01, 0.05, 0.1, 1\}$.}
  \label{fig:params}
\end{figure}

\subsection{Parameter Sensitivity Analysis}

In DIMVC-HIA, two trade-off hyperparameters, $\alpha$ and $\beta$, are introduced to balance multiple objectives in the loss function. To assess the model’s sensitivity to these parameters, we conducted experiments on the BDGP and Handwritten datasets under a missing ratio ($\eta$) of 0.3. As shown in Figure~\ref{fig:params}, the clustering performance (measured by ACC) remains consistently strong as $\alpha$ varies in the range of $[0.01, 0.10]$ and $\beta$ in $[0.01, 0.05]$. The results indicate that the overall performance is relatively stable, exhibiting no substantial fluctuations across the tested configurations. Consequently, we choose $\alpha = 0.1$ and $\beta = 0.01$ as the default settings for all datasets.

\section{Conclusion}
This paper presents DIMVC-HIA, a hierarchical imputation and alignment framework for incomplete multi-view clustering, which integrates view-specific autoencoders, energy-based alignment, and contrastive cluster assignment to ensure reliable imputation, semantic consistency, and compact clustering under varying levels of view incompleteness. Extensive experiments demonstrate DIMVC-HIA’s superiority over IMVC baseline methods across a range of missing ratios, with results from multiple evaluation metrics confirming its ability to produce well-structured and discriminative representations under severe view incompleteness.

\section*{Acknowledgements}
This work was supported in part by the NSF (CNS-2153358, CNS-2245918), the DoD CoE-AIML (W911NF-20-2-0277), and the Commonwealth Cyber Initiative (CCI and COVA CCI).

\bibliography{aaai2026}

\end{document}